\documentclass[conference]{IEEEtran}
\IEEEoverridecommandlockouts

\usepackage{cite}
\usepackage{amsmath,amssymb,amsfonts}
\usepackage{mathtools}
\usepackage{xcolor}
\usepackage{url}

\DeclareMathOperator*{\argmin}{arg\,min}

\def\BibTeX{{\rm B\kern-.05em{\sc i\kern-.025em b}\kern-.08em
    T\kern-.1667em\lower.7ex\hbox{E}\kern-.125emX}}
\begin{document}

\title{Take Goodhart Seriously: Principled Limit on General-Purpose AI Optimization}

\author{
\IEEEauthorblockN{Antoine Maier\IEEEauthorrefmark{1}\thanks{Corresponding author: amaier@gpaipolicylab.org}
                  \quad Aude Maier\IEEEauthorrefmark{2}
                  \quad Tom David\IEEEauthorrefmark{1}}
\IEEEauthorblockA{\IEEEauthorrefmark{1}General-Purpose AI Policy Lab (GPAI Policy Lab), Paris, France}
\IEEEauthorblockA{\IEEEauthorrefmark{2}École polytechnique fédérale de Lausanne (EPFL), Lausanne, Switzerland}
}

\maketitle

\begin{abstract}
A common but rarely examined assumption in machine learning is that training yields models that actually satisfy their specified objective function. We call this the \textit{Objective Satisfaction Assumption} (OSA). Although deviations from OSA are acknowledged, their implications are overlooked. We argue, in a learning-paradigm-agnostic framework, that OSA fails in realistic conditions: approximation, estimation, and optimization errors guarantee systematic deviations from the intended objective, regardless of the quality of its specification. Beyond these technical limitations, perfectly capturing and translating the developer's intent, such as alignment with human preferences, into a formal objective is practically impossible, making misspecification inevitable. Building on recent mathematical results, absent a mathematical characterization of these gaps, they are indistinguishable from those that collapse into Goodhart’s law failure modes under strong optimization pressure. Because the Goodhart breaking point cannot be located ex ante, a principled limit on the optimization of General-Purpose AI systems is necessary. Absent such a limit, continued optimization is liable to push systems into predictable and irreversible loss of control. 
\end{abstract}


\section{Introduction}

In the machine learning community, reinforcement learning (RL) is commonly described as producing agents whose goal is to maximize the expected cumulative reward, or return. This formulation even appears in \textit{Reinforcement Learning: An Introduction}\cite{sutton_reinforcement_2020}, widely regarded as the most influential RL introductory textbook, whose authors received the 2024 Turing Award for ``developing the conceptual and algorithmic foundations of reinforcement learning\cite{association_for_computing_machinery_acm_2025}.'' This view is not specific to RL, but reflects a broader and often overlooked assumption in machine learning: the training results in a model that satisfies the specified learning objective. In other words, it is commonly believed that if a model is trained to minimize a loss function or maximize a reward signal, the resulting model will indeed act in a way that optimizes that objective. We refer to this belief as the \textit{Objective Satisfaction Assumption (OSA)}. Yet it is well acknowledged that approximation, estimation, and optimization errors prevent models from perfectly satisfying their objective\cite{bottou_tradeoffs_2007}. Taken seriously, this implies that we cannot robustly implement the intended goal in an AI system, regardless of the quality of its specification.

But the challenge begins even before training. In most real-world applications, the specified objective function is only a proxy for the developer's true intent\cite{amodei_concrete_2016}. This \textit{specification problem} arises because high-level human goals are difficult to express precisely in a formal mathematical framework. 

The combination of these two issues, the failure of the OSA and the specification problem, results in a systematic discrepancy between the developer's intent and the actual behavior of the trained model. Although the model may perform approximately as intended in many situations, recent mathematical results\cite{el-mhamdi_goodharts_2024,majka_strong_2025} show that under strong optimization pressure, this discrepancy can lead to significant failures, a phenomenon known as Goodhart's law\cite{goodhart_problems_1984}. Under strong optimization, a strong Goodhart regime, where the developer's utility becomes negatively correlated with the proxy and can even decrease without bound, plausibly represents the default. Since the discrepancy is unlikely to satisfy the narrow conditions required for more favorable outcomes, and the Goodhart breakpoint cannot be located ex ante, continued optimization without a principled limit risks crossing into a regime where the model's behavior becomes progressively opposed to what was intended.

This is particularly concerning in the context of General-Purpose AI (GPAI) systems, which are designed to ``satisfy human preferences\cite{russell_human_2019,bai_constitutional_2022}'' and are developed to perform a wide range of tasks and adapt to various situations. Since misspecification and OSA failure prevent both the accurate capture of the developer's intent in the learning objective and its reliable implementation in the trained model, GPAI systems satisfy the conditions for Goodhart failure where the actions actually taken by the system diverge from what was intended by the developer. As these systems are optimized for greater performance, generality and autonomy, optimization pressures favor the emergence of behaviors that expand the set of reachable high-performance trajectories, such as functioning preservation and capability improvements, known as \textit{convergent instrumental goals}\cite{omohundro_basic_2008,bostrom_superintelligent_2012}. Conversely, control mechanisms that restrict the model's autonomy suppress high-performance trajectories, making resistance to such mechanisms a systematically favored strategy under optimization. Because of misspecification and OSA failure, selection pressures that would counteract instrumental convergence cannot be reliably specified or implemented. The expected Goodhart failure mode of agentic GPAI under strong optimization pressure is therefore for such systems to pursue misaligned goals while resisting correction, both reactively and preemptively, as their capacity to undermine human oversight grows, leading to irreversible loss-of-control scenarios.

These scenarios were first anticipated conceptually\cite{omohundro_basic_2008,bostrom_superintelligent_2012}, then formalized\cite{benson-tilsen_formalizing_2016,turner_optimal_2023}, and precursors are now observed empirically\cite{meinke_frontier_2025,greenblatt_alignment_2024,lynch_agentic_2025,weij_ai_2025,bondarenko_demonstrating_2025}. Given the irreversibility of loss of control, the impossibility of locating the Goodhart breakpoint ex ante, and the incentive for capable agents to preemptively circumvent oversight mechanisms, a principled limit on GPAI optimization is necessary.

\begin{figure}
    \centering
    \includegraphics[width=0.75\linewidth]{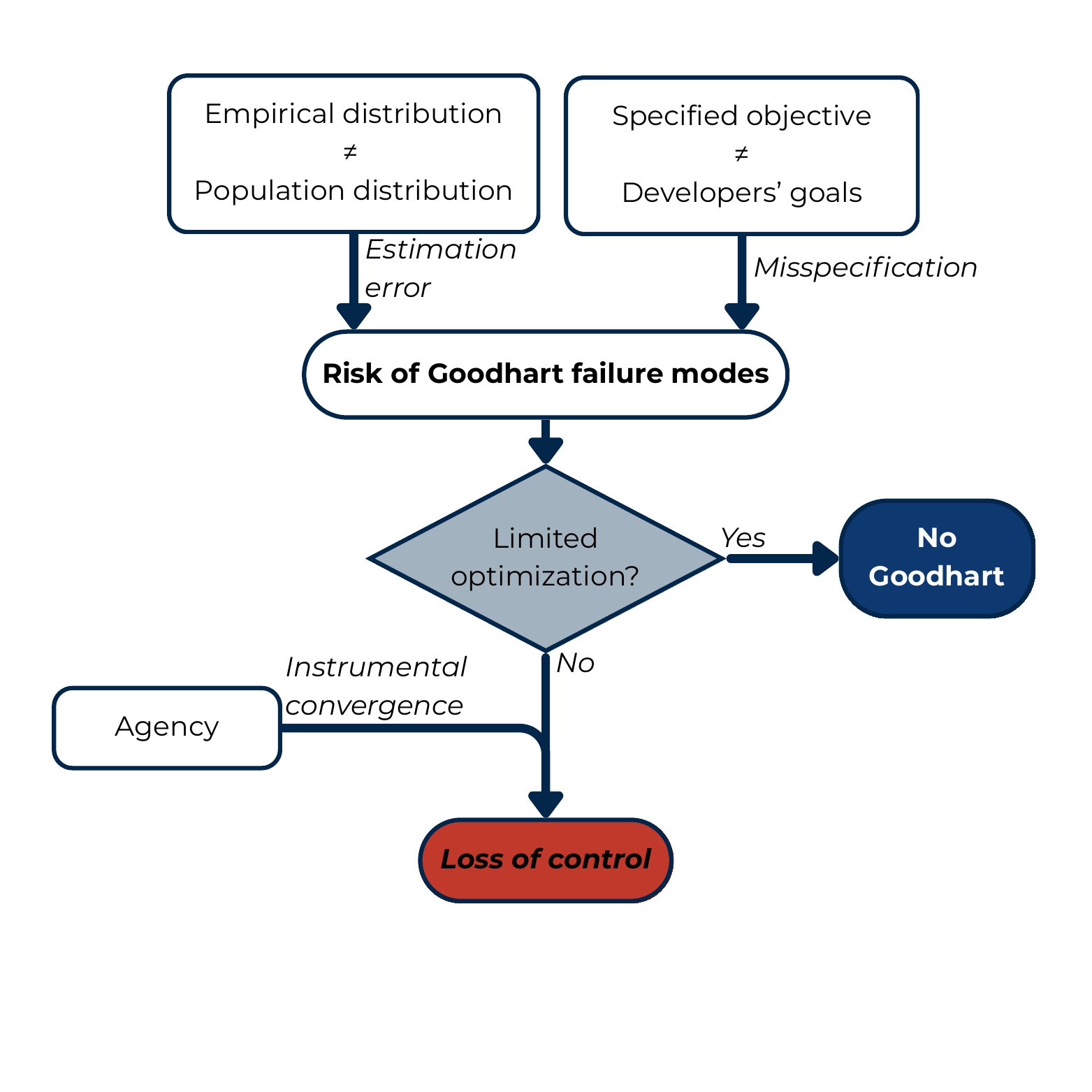}
    \caption{Schematic overview of the argument. Estimation error and misspecification each introduce a discrepancy between the developer's intent and the model's learned objective, creating the conditions for Goodhart failure. If optimization is limited, the risk is contained. Otherwise, instrumental convergence due to agency of General-Purpose AI models leads to loss of control.}
    \label{fig:schema}
\end{figure}

Our reasoning is represented schematically in Fig.~\ref{fig:schema}.

In essence, this paper constructs a unified argument chain connecting well-acknowledged machine learning limitations to concrete safety risks, grounding common intuitions in AI safety in formal and empirical results. While the individual building blocks, approximation--estimation--optimization errors, misspecification, Goodhart's law, and instrumental convergence, are each discussed in their respective literatures, they are rarely connected into a single coherent argument. Our main contributions are as follows:
\begin{enumerate}
    \item We introduce the \textit{Objective Satisfaction Assumption} (OSA), naming and formalizing the often-implicit belief that training yields models satisfying their specified objective. By recasting well-known approximation, estimation, and optimization errors within a recent learning-paradigm-agnostic framework\cite{hu_toward_2023}, we make explicit that this assumption fails in all realistic settings;
    \item We connect the OSA failure and the well-known specification problem to recent formal results on Goodhart's law\cite{el-mhamdi_goodharts_2024,majka_strong_2025}, arguing that the inevitable discrepancy between proxy and intended goal places GPAI development in a candidate setting for strong Goodhart failure;
    \item We construct an explicit argument chain: conditioned on instrumental convergence of agentic GPAI, Goodhart premises lead to loss-of-control scenarios under strong optimization. We conclude that a principled limit on GPAI optimization is therefore necessary.
\end{enumerate}

The remainder of this paper is organized as follows. Sec.~\ref{sec:related-work} reviews related work. Sec.~\ref{sec:OSA} defines the OSA and argues it fails in realistic conditions. Sec.~\ref{sec:misspecification} discusses the specification problem. Sec.~\ref{sec:consequences} connects these issues to Goodhart's law and argues that, for increasingly agentic GPAI, they give rise to loss-of-control scenarios calling for a principled limit on optimization. Sec.~\ref{sec:conclusion} concludes.

\section{Related Work}
\label{sec:related-work}

The unified learning framework we use is from Hu et al.\cite{hu_toward_2023}, who proposed the \textit{Standard Equation of Machine Learning} to encompass many learning paradigms.

The Objective Satisfaction Assumption (OSA) is not explicitly named in the literature, but the limitations of learning algorithms are well-known. The decomposition of the excess risk into approximation, estimation, and optimization errors is a standard framework in statistical learning theory\cite{bottou_tradeoffs_2007,shalev-shwartz_understanding_2014}. The term \textit{inner misalignment} has been used to describe the discrepancy between the learned model and the true optimal model due to these errors\cite{hubinger_risks_2021}.

Separately, the specification problem, or \textit{outer misalignment}, has been extensively discussed in the AI safety literature\cite{amodei_concrete_2016,hubinger_risks_2021}. The difficulty of specifying human values and preferences into a formal objective function is a central challenge in the field.

Goodhart's law was first proposed in the context of economics\cite{goodhart_problems_1984}, but its implications for AI alignment have been explored in various works\cite{manheim_categorizing_2019,shah_goal_2022}, including recent rigorous mathematical formalizations of this phenomenon\cite{el-mhamdi_goodharts_2024, majka_strong_2025, zhuang_consequences_2021}.

Loss-of-control scenarios have been anticipated conceptually\cite{omohundro_basic_2008,bostrom_superintelligent_2012}, formalized mathematically\cite{benson-tilsen_formalizing_2016,turner_optimal_2023}, and precursors are starting to be observed empirically\cite{meinke_frontier_2025,greenblatt_alignment_2024,lynch_agentic_2025,weij_ai_2025,bondarenko_demonstrating_2025}.

Finally, a limited body of work directly engages with critiques of the loss-of-control argument. Some dispute the inevitability of instrumental convergence\cite{thorstad_what_2024,southan_timing_2025}, while others survey popular counterarguments to loss-of-control scenarios, ultimately finding them unconvincing\cite{swoboda_examining_2025}.

\section{Objective Satisfaction Assumption}
\label{sec:OSA}

In this section, we define the paradigm-agnostic learning framework and the Objective Satisfaction Assumption (OSA), and then argue that it is almost surely false in realistic conditions.

\subsection{Learning Framework}

Hu et al.\cite{hu_toward_2023} recently proposed a unified framework encompassing many, if not all, learning paradigms, centered on the \textit{Standard Equation of Machine Learning}, defining a learning problem as the minimization of an objective function over a hypothesis class. We list its main components below, hiding irrelevant details, and present the resulting equation (Eq.~\ref{eq:empirical-learning-problem}).

\paragraph{Objective function}
Regardless of the specific machine learning paradigm (supervised, unsupervised, self-supervised, RL, etc.), a learning task is an optimization problem where an \textit{objective function} $\mathcal{L}$ is extremized by a learning algorithm. The objective function is composed of an \textit{experience term} $\mathcal{E}$, which provides feedback to the learning algorithm on the model's performance, and \textit{regularization terms} $\Omega$, which guide the algorithm toward well-behaved solutions. These components are detailed in the paragraphs below. Without loss of generality, if $\mathcal{L}$ is to be minimized, any monotone transform can be maximized instead, e.g.\ $\mathcal{M} \coloneq \exp(-\mathcal{L})$. We use minimization and maximization interchangeably, since they are equivalent up to this transformation.

\paragraph{Experience term}
The main driver of the learning process is the \textit{experience term} $\mathcal{E}$, which captures feedback from the environment, typically a loss measuring prediction error in supervised learning, a reward signal in RL, or a reconstruction error or likelihood in unsupervised learning.

\paragraph{Population vs empirical distribution}
Machine learning aims to build models that perform well on the distribution of real-world scenarios encountered after deployment, the \textit{population distribution} $\mathbb{P}$. Since this distribution is inaccessible, models are trained on an approximation, e.g., a finite dataset or a simulated environment, called the \textit{empirical distribution} $\widetilde{\mathbb{P}}$. This yields two forms of the experience term: the population $\mathcal{E}$, evaluated over $\mathbb{P}$, and the empirical $\widetilde{\mathcal{E}}$, evaluated over $\widetilde{\mathbb{P}}$ (typically replacing expectations by averages). Correspondingly, we write $\mathcal{L}$ for the population objective and $\widetilde{\mathcal{L}}$ for its empirical counterpart.

\paragraph{Hypothesis class}
The optimization is performed over a \textit{hypothesis class} $\mathcal{F}$, which is the set of functions that the learning algorithm can choose from. If the model $f_\theta$ is parameterized by a set of parameters $\theta \in \Theta$, then $\mathcal{F} = \mathcal{F}_\Theta$ corresponds to the set of functions that can be represented by varying these parameters. For example, in neural networks, for a fixed architecture, $\mathcal{F}_\Theta$ would be the set of all functions that can be represented by that architecture with different weight values $\theta \in \Theta$. More generally, $\Theta$ can include different architectures and non-neural parameterizations. We define $\mathcal{F}_\infty$ as the space of all functions, independent of parameterization, so $\mathcal{F}_\Theta \subset \mathcal{F}_\infty$.

\paragraph{Regularization terms}
The empirical experience term is often complemented by \textit{regularization terms} that favor simpler or more robust solutions. Common examples include L1/L2 penalties and exploration bonuses in RL. Regularization is needed due to limited data and compute; with unlimited resources, the experience term alone would suffice. Regularization terms can depend on the parameters $\theta$ or on the function $f_\theta$ itself. We write $\Omega(f_\theta)$ for their combined contribution.

Putting all these components together, we can write the machine learning problem in a general form:
\begin{equation}
\label{eq:empirical-learning-problem}
    \min_{\theta \in \Theta} \widetilde{\mathcal{L}}(f_\theta) \coloneq \widetilde{\mathcal{E}}(f_\theta) + \lambda \Omega(f_\theta)
\end{equation}
where $\lambda$ is a hyperparameter that controls the strength of the regularization relative to the experience term. If we had access to the true population distribution and were free from representational capacity or other computational constraints, regularization would be unnecessary, and the learning problem Eq.~\ref{eq:empirical-learning-problem} would become:
\begin{equation}
\label{eq:population-learning-problem}
    \min_{f \in \mathcal{F}_\infty} \mathcal{L}(f) \coloneq \mathcal{E}(f).
\end{equation}

If we write $\hat{\theta}$, or equivalently $f_{\hat{\theta}}$, for the actual output of the optimization algorithm, then the \textit{Objective Satisfaction Assumption} (OSA) states that $f_{\hat{\theta}}$ is a solution of the population learning problem, i.e., 
\begin{equation}
\label{eq:OSA}
    f_{\hat{\theta}} \in \argmin_{f \in \mathcal{F}_\infty} \mathcal{L}(f)
\end{equation}
where we used $\in$ instead of $=$ to account for the possibility of multiple solutions to the population learning problem. 

\subsection{Why the Objective Satisfaction Assumption Fails}
\label{sec:inner-misalignment}

In an idealized scenario with unrestricted representational capacity $\mathcal{F}_\Theta = \mathcal{F}_\infty$, perfect distribution $\widetilde{\mathbb{P}} = \mathbb{P}$, and unlimited compute to find the global minimum, the OSA (Eq.~\ref{eq:OSA}) would hold\cite{vapnik_principles_1991} and the experience term $\mathcal{E}$ would entirely determine the model's behavior (in the limit $\lambda \to 0$). In practice, various limitations lead to the failure of the OSA.

It is usual\cite{bottou_tradeoffs_2007} to define:
\begin{equation}
\begin{aligned}
    & f^*\in \argmin_{f\in\mathcal{F}_\infty} \mathcal{L}(f) && \text{(Bayes/unrestricted optimum)},\\
    & \theta^*\in\argmin_{\theta\in\Theta} \mathcal{L}(f_\theta) && \text{(best in class)},\\
    & \widetilde{\theta}^*\in\argmin_{\theta\in\Theta} \widetilde{\mathcal{L}}(f_\theta) && \text{(Empirical Risk Minimization in class)},\\
    & \hat{\theta}\in\Theta && \text{(output of the optimization algorithm)}.
\end{aligned}
\end{equation}
to then decompose the excess risk as follows:
\begin{equation}
\label{eq:excess-risk-L}
\begin{aligned}
    & \mathcal{L}(f_{\hat{\theta}}) - \mathcal{L}(f^*) \\
    &= \underbrace{\mathcal{L}(f_{\hat{\theta}}) - \mathcal{L}(f_{\widetilde{\theta}^*})}_{\text{Optimization error\,} \geq 0}
    + \underbrace{\mathcal{L}(f_{\widetilde{\theta}^*}) - \mathcal{L}(f_{\theta^*})}_{\text{Estimation error\,} \geq 0}
    + \underbrace{\mathcal{L}(f_{\theta^*}) - \mathcal{L}(f^*)}_{\text{Approximation error\,} \geq 0} \\
    &\geq 0
\end{aligned}
\end{equation}
where all terms are evaluated on the population learning objective. Note that the three error terms are nonnegative and the OSA (Eq.~\ref{eq:OSA}) holds if and only if all three are zero. This decomposition allows us to identify distinct error sources.

\paragraph{Approximation error}
This error is due to the limited representational capacity of $\mathcal{F}_\Theta \subset \mathcal{F}_\infty$\cite{cucker_learning_2007}. While no Universal Approximation Theorem yet encompasses all modern architectures, existing results\cite{hornik_multilayer_1989,leshno_multilayer_1993,zhou_universality_2020,yun_are_2020} support the conjecture that sufficiently expressive neural networks can approximate a broad class of functions to arbitrary precision. However, this holds only in the limit of infinitely large models, so in practice a nonzero approximation error remains.

\paragraph{Estimation error}
This error arises because learning uses the empirical distribution $\widetilde{\mathbb{P}}$ rather than the population $\mathbb{P}$\cite{shalev-shwartz_understanding_2014}. Although the Law of Large Numbers guarantees convergence of empirical estimates to population quantities in the infinitely many i.i.d. samples limit\cite{vapnik_nature_2000}, real datasets are finite, often biased, and non-i.i.d.\cite{quinonero-candela_dataset_2022}, further accentuated by data augmentation. 

\paragraph{Optimization error}
This error arises because the optimization algorithm outputs only an approximate solution $\hat{\theta}$ rather than $\widetilde{\theta}^*$\cite{bottou_tradeoffs_2007}. While iterative methods can in principle converge to a global minimum given infinite resources\cite{hajek_cooling_1988}, limited compute constrains learning dynamics\cite{blum_training_1988}, and a nonzero optimization error remains, often deliberately (e.g., early stopping).

Any single nonzero error term suffices to falsify the OSA, and in practice all are nonzero. Therefore, the trained model almost surely does not satisfy the objective function (e.g. misclassified images must exist in computer vision, suboptimal state-action pairs must exist in RL), regardless of how well specified the learning objective is. 

The difference between the Bayes-optimal solution $f^*$ and the actual solution resulting from the training $f_{\hat{\theta}}$ that arises even in scenarios where the learning objective is perfectly specified is sometimes referred to as \textit{inner misalignment}\cite{hubinger_risks_2021}.

\subsection{Misgeneralization}
\label{sec:misgeneralization}

While approximation and optimization errors limit access to the Bayes optimum due to computational limitations, they leave the learning objective itself unchanged. On the other hand, finite-sample estimation of the population distribution is of greater consequence on the learning dynamics since it directly alters the learning objective, i.e., $\widetilde{\mathcal{L}} \neq \mathcal{L}$.

Since the empirical distribution $\widetilde{\mathbb{P}}$ is only an approximation of the population distribution $\mathbb{P}$, it can differ in various ways. It can contain spuriously correlated features and overweight or omit genuinely relevant others. A model trained on $\widetilde{\mathbb{P}}$ will perform well on inputs it has effectively seen, and often generalize well on new inputs that share similar underlying structure (the “magic” of deep learning). On this subset of the input space, we typically have $f_{\hat{\theta}} \approx f^*$.

However, when faced with inputs that are out of (empirical) distribution, either because formerly correlated features are disentangled, or because a salient feature never present in $\widetilde{\mathbb{P}}$ suddenly appears, the learned function will, in general, misgeneralize\cite{damour_underspecification_2022}. For example, a vision model trained to recognize insects on natural images may fail when insects appear on atypical backgrounds, because it has learned to associate foliage backgrounds with the insect class\cite{yang_identifying_2024}. Empirically, these failure modes are common\cite{damour_underspecification_2022,shah_goal_2022} and they can appear on human-level concepts as the previous example illustrates, but also on abstract latent features\cite{olah_feature_2017,bau_network_2017}.

Even more so, deep learning has a simplicity bias\cite{valle-perez_deep_2019, soudry_implicit_2018} that actively favors the search and exploitation of these correlated features, referred to as \textit{shortcut learning}\cite{geirhos_shortcut_2020}, to free representational budget for other features\cite{elhage_toy_2022}. Under optimization pressure, the same bias favors the spontaneous emergence of \textit{mechanistic generalization circuits}, such as \textit{internal world models}\cite{li_emergent_2023} or \textit{in-context learning} via inference-time optimization algorithms\cite{akyurek_what_2023, bai_transformers_2023, von_oswald_transformers_2023}, which generally improve generalization\cite{olsson_-context_2022}. 

Yet this success has limits. No finite sample distribution $\widetilde{\mathbb{P}}$ can fully recover the true distribution $\mathbb{P}$. If the inductive bias aligns with the structures of the world, the model correctly generalizes. But there are vastly more predictors consistent on $\widetilde{\mathbb{P}}$ than there are Bayes-optimal ones, since the finite sampling leaves the learning problem underspecified. Therefore, the generalization mechanisms can themselves be spurious, and inputs that elicit misgeneralization will persist even as models and datasets scale\cite{shah_goal_2022}. 

\subsection{Performativity}

A growing area of research highlights another error source that arises from the interaction between the learning system and its environment. Classical machine learning assumes the learning problem is external to the environment, reflected in the static nature of the objective function. This ignores that a model can influence the environment it evolves in, thereby changing the data distribution. This is known as performativity\cite{perdomo_performative_2020}. A canonical example is a recommendation system whose suggestions change user behavior over time. The greater the system's transformative power, the more significant this retroactive effect\cite{hardt_performative_2022}.

In the static framework, the learning objective Eq.~\ref{eq:empirical-learning-problem} defines a static \textit{performance landscape} over the parameter space $\Theta$ where regions with low values of $\widetilde{\mathcal{L}}$ correspond to models $f_\theta$ with good performance. The role of the optimization algorithm is to explore this landscape, and it aims to find the region of best performance.

In the performative setting, however, the objective function itself depends on the current parameters $\theta$, i.e., $\widetilde{\mathcal{L}}(\theta)$. Although the optimization algorithm still favors regions of low $\widetilde{\mathcal{L}}$, these regions change as the parameters change. Therefore, the performance landscape is dynamic and changes as the optimization algorithm explores it.

This retroactive loop may unfold across retraining cycles, but it can also arise within a single training phase, for example in continual learning, or even during pretraining if the system has enough performative power. As future GPAI become more transformative (through general adoption, multimodality, performance, etc.), such feedback effects will likely intensify.

For simplicity, we stay in the static framework. Although performativity in machine learning is an active research area\cite{hardt_performative_2025}, the results are too preliminary to incorporate into our argument. Nevertheless, it adds another layer of complexity, further reinforcing the conclusion that the resulting model deviates from the OSA.

\section{Misspecification}
\label{sec:misspecification}

Even if the OSA were to hold (Eq.~\ref{eq:OSA}), a second type of error would remain: the learning objective function is misspecified relative to the developer's implicitly intended goal\cite{amodei_concrete_2016}. This \textit{specification problem} arises because high-level human goals are difficult to express precisely in a formal mathematical framework. This is sometimes referred to as \textit{outer misalignment}\cite{hubinger_risks_2021}. 

To isolate the specification problem, assume the ideal scenario where the OSA holds: consider the population objective $\mathcal{M} (\coloneq \exp(-\mathcal{L}))$ on the unrestricted hypothesis class $\mathcal{F}_\infty$, ignoring computational limitations. 

Framing learning as an optimization problem (Eq.~\ref{eq:population-learning-problem}) implicitly assumes the existence of a utility function that the developer wants to maximize\cite{russell_artificial_2016}. Whereas $\mathcal{M}: \mathcal{F}_\infty \to \mathbb{R}$ is a utility function that measures the quality of any candidate model $f \in \mathcal{F}_\infty$, it is a proxy for the true goal $\mathcal{G}: \mathcal{F}_\infty \to \mathbb{R}$, which captures how well $f$ satisfies the developer's intent. Not only would $\mathcal{G}$ assign high values to models that perform well at the given task, it also captures all human-level desiderata. A model considered truthful, fair, and helpful for the developer would have a higher $\mathcal{G}$ than one that is not.

The learning algorithm requires the objective to be an explicit mathematical function. While $\mathcal{G}$ can in principle be implicitly defined, human preferences are vague, context-dependent, and leave many edge cases implicit~\cite{abbeel_apprenticeship_2004}, making exhaustive specification infeasible. The proxy $\mathcal{M}$ fulfills this need, but translating implicit preferences into an explicit function is a historically difficult problem that remains unsolved. While $\mathcal{M}$ is hopefully correlated with $\mathcal{G}$, it cannot capture it fully, and the misspecification $\mathcal{M} \neq \mathcal{G}$ is inevitable. As a result, any eventual bound on the excess error Eq.~\ref{eq:excess-risk-L} does not translate to $\mathcal{G}$ and the true implicit excess error
\begin{equation}
    \mathcal{G}(f_{\hat{\theta}}) - \sup_{f \in \mathcal{F}_\infty} \mathcal{G}(f)
    \neq 
    \mathcal{L}(f_{\hat{\theta}}) - \mathcal{L}(f^*)
\end{equation}
can a priori be arbitrarily large even when $\hat{\theta}$ is (near-)optimal for $\mathcal{L}$.

Current approaches such as RLHF learn a reward model $\mathcal{M}_\text{learned}$ from human or AI feedback as a proxy for human preferences $\mathcal{G}$, and use it to fine-tune GPAI\cite{christiano_deep_2017, stiennon_learning_2020, ouyang_training_2022, bai_constitutional_2022}. But the reward model itself is subject to the same limitations (Sec.~\ref{sec:inner-misalignment}), so a misspecification $\mathcal{M}_\text{learned} \neq \mathcal{G}$ remains.

To compensate for the difficulty of specifying a single objective function that captures the learning objective fully, modern GPAI undergo multi-stage training (pretraining, RLHF, safety fine-tuning, etc.), each stage with a different objective and regularizations imposing proximity to the previous stage\cite{stiennon_learning_2020, ouyang_training_2022}. While this helps match frontier labs' product design targets, it creates a blend of partially conflicting objectives\cite{casper_open_2023}, adding another layer of difficulty to establishing safety guarantees.

The purpose of GPAI is often framed as satisfying human preferences\cite{russell_human_2019, bai_constitutional_2022}, but this overlooks two fundamental difficulties. First, even widely valued concepts such as truthfulness or ethical behavior are debated and interpreted differently across individuals and cultures\cite{jobin_global_2019}. Possible avenues for a solution have been proposed such as determining a consensus objective through a democratic process\cite{ovadya_toward_2024}. Second, even for a single individual, a coherent implicit utility function $\mathcal{G}$ rarely exists\cite{kahneman_prospect_2013}. Empirical experiments show systematic inconsistencies with the von Neumann-Morgenstern axioms\cite{von_neumann_theory_2007}. Alternatives such as Coherent Extrapolated Volition\cite{yudkowsky_coherent_2004} have been proposed to capture what individuals truly value, but specifying this notion is no easier. In the absence of a valid expected-utility framework, any purely maximization-based approach is ultimately bound to fail.

\section{Consequences in AI Safety}
\label{sec:consequences}

The preceding sections argued that the trained system $f_{\hat{\theta}}$ does not satisfy the learning objective $\mathcal{L}$ (Sec.~\ref{sec:inner-misalignment}), and $\mathcal{L}$ (or equivalently $\mathcal{M}$) is, at best, a proxy for the true objective $\mathcal{G}$ (Sec.~\ref{sec:misspecification}). We now examine how these gaps imply concrete safety risks.

\subsection{Goodhart's Law under High Optimization Pressure}
\label{sec:goodhart}

We want to study how the correlation between the proxy objective $\mathcal{M}$ and the true goal $\mathcal{G}$ evolves under strong optimization pressure, and in particular when the correlation breaks. This phenomenon is known as \textit{Goodhart's law}\cite{goodhart_problems_1984}: When a measure becomes a target, it ceases to be a good measure\cite{strathern_improving_1997}.

Note that the estimation error that leads to misgeneralization (Sec.~\ref{sec:misgeneralization}) can be seen as a second source of misspecification, where the finite empirical distribution $\widetilde{\mathbb{P}}$ used to estimate the population distribution $\mathbb{P}$ specifies the empirical learning objective $\widetilde{\mathcal{M}}$ only on a subset of the input space. The present discussion will focus on the discrepancy between $\widetilde{\mathcal{M}}$ and $\mathcal{G}$, but the same reasoning applies when considering the discrepancy between the population $\mathcal{M}$ and $\mathcal{G}$, or even between $\widetilde{\mathcal{M}}$ and $\mathcal{M}$. 

The training of the learning objective Eq.~\ref{eq:empirical-learning-problem} defines a probability measure over the parameter space $\Theta$, denoted $\hat{\mathbb{P}}_\Theta$. For example, minimizing the loss with stochastic gradient descent with random initialization on a fixed training set and a stopping criterion defines a distribution where the regions of $\Theta$ that yield lower values of $\widetilde{\mathcal{L}}$ and that the optimization algorithm can reach typically have higher probability. This distribution is in general complex and depends on many factors, such as the learning objective $\widetilde{\mathcal{L}}$, because it drives the optimization dynamics, but also on the optimization algorithm itself (its initialization, its hyperparameters, its stopping criterion, etc.), on the model architecture (which defines the hypothesis class $\mathcal{F}_\Theta$), among other factors.

We consider another instance of the learning objective $\widetilde{\mathcal{M}}$, together with the true goal $\mathcal{G}$. Both are utility functions $\widetilde{\mathcal{M}}, \mathcal{G}: \mathcal{F}_\Theta \to \mathbb{R}$ assigning a score to any candidate model $f_\theta \in \mathcal{F}_\Theta$. Through the probability measure $\hat{\mathbb{P}}_\Theta$ on $\Theta$, these functions induce random variables $\widetilde{\mathcal{M}}(\theta)$ and $\mathcal{G}(\theta)$. For simplicity, we will henceforth use the same symbols $\widetilde{\mathcal{M}}$ and $\mathcal{G}$ to denote these random variables.

As the learning task is optimized, the probability mass of the distribution $\hat{\mathbb{P}}_\Theta$ concentrates more and more on regions of $\Theta$ where $\widetilde{\mathcal{M}}$ is high. To analyze the behavior in this limit, we consider the conditional distribution 
\begin{equation}
    \hat{\mathbb{P}}_{\Theta | \alpha} \coloneq
    \hat{\mathbb{P}}_\Theta\left(\cdot \middle| \widetilde{\mathcal{M}} \geq q_\alpha(\widetilde{\mathcal{M}})\right)
\end{equation}
where $q_\alpha(\widetilde{\mathcal{M}})$ is the $\alpha$-quantile of $\widetilde{\mathcal{M}}$. As $\alpha \to 0$, this conditional distribution concentrates on regions of $\Theta$ where $\widetilde{\mathcal{M}}$ is high, corresponding to strong optimization pressure.

El-Mhamdi and Hoang\cite{el-mhamdi_goodharts_2024} and Majka and El-Mhamdi\cite{majka_strong_2025} recently studied the asymptotic behavior of the correlation between $\widetilde{\mathcal{M}}$ and $\mathcal{G}$,
\begin{equation}
    \rho_\alpha \coloneq
    \mathrm{corr}\left( \widetilde{\mathcal{M}}, \mathcal{G} \middle| \widetilde{\mathcal{M}} \geq q_\alpha(\widetilde{\mathcal{M}}) \right)
\end{equation}
in the limit of strong optimization pressure $\alpha \to 0$. They showed that the Goodhart regime depends critically on the tail behavior of the discrepancy
\begin{equation}
    \xi \coloneq \widetilde{\mathcal{M}} - \mathcal{G}
\end{equation}
relative to that of $\mathcal{G}$. Across the distributional settings studied, the results suggest that when $\mathcal{G}$ has a heavier tail than $\xi$, optimization remains useful, a situation termed \textit{benign Goodhart}\cite{majka_strong_2025}. As the tail of $\xi$ becomes heavier relative to $\mathcal{G}$, however, the picture progressively worsens, from a \textit{weak Goodhart} regime where $\rho_\alpha \to 0^+$ and optimization becomes useless, to a \textit{strong Goodhart} regime where $\rho_\alpha$ becomes negative and the expected goal $\mathbb{E}[\mathcal{G} | \widetilde{\mathcal{M}} \geq q_\alpha]$ decreases under further optimization, meaning that optimizing the proxy becomes actively harmful. The strong regime can additionally be \textit{catastrophic}, with $\mathbb{E}[\mathcal{G} | \widetilde{\mathcal{M}} \geq q_\alpha] \to -\infty$\cite{el-mhamdi_goodharts_2024}. While these results are proven for specific distributional families, they suggest that the relative tail thickness is the primary determinant of the regime. Intuitively, as $\widetilde{\mathcal{M}}$ is over-optimized, the selected region of the parameter space is increasingly dominated by models where the discrepancy $\xi$ is large, and the goal $\mathcal{G}$ is dwarfed in comparison.

This formalizes Goodhart's law. When the proxy objective $\widetilde{\mathcal{M}}$ misspecifies the true goal $\mathcal{G}$, and omits goal-relevant features, optimization pressures trade off arbitrarily large degradation of these unconstrained features for arbitrarily small gains in the proxy. We saw in Sec.~\ref{sec:misspecification} and Sec.~\ref{sec:misgeneralization} that such misspecification is inevitable in practice. Cases of \textit{specification gaming}\cite{krakovna_specification_2020}, where the system finds loopholes to superficially satisfy the proxy objective without satisfying the intended goal, are manifestations of Goodhart's law. The canonical illustration is the CoastRunners agent, which learns to loop endlessly to collect checkpoint rewards instead of finishing the race, thereby maximizing $\widetilde{\mathcal{M}}$ while failing at $\mathcal{G}$\cite{amodei_concrete_2016}. But GPAI, trained to follow the user's instructions, exhibit similar behavior. A recent example shows that when instructed to win against a chess engine, state-of-the-art reasoning models like o3 or DeepSeek-R1 often exploit flaws in the chess environment to register wins, once they infer the opponent is too strong to defeat through legal play\cite{bondarenko_demonstrating_2025}.

In practice, however, one cannot determine in advance which regime applies to a given learning problem, since the true goal $\mathcal{G}$ is inexpressible in mathematical terms, which is precisely the specification problem discussed in Sec.~\ref{sec:misspecification}. Consequently, the tail distribution of $\xi$ is equally inaccessible. Avoiding strong Goodhart requires $\xi$ to have a thinner tail than $\mathcal{G}$, but for GPAI systems designed to satisfy complex, context-dependent human preferences, the discrepancy can grow along any of the many dimensions left unconstrained by the proxy (Sec.~\ref{sec:misgeneralization}), making this favorable condition unlikely to hold. The strong Goodhart regime therefore represents a plausible default for GPAI under continued optimization. Since the Goodhart breakpoint cannot be located ex ante, a principled limit on optimization pressure is necessary.

\subsection{Loss of Control under High Optimization Pressure}

General-Purpose AI systems are designed to perform a wide range of tasks and adapt to diverse situations\cite{russell_human_2019, bai_constitutional_2022}. Frontier labs are actively pursuing increasingly agentic systems\cite{openai_new_2025, anthropic_ai_nodate, google_deepmind_project_nodate}, driven by their unprecedented economic, scientific, and military potential\cite{schmidt_final_2021}. 

The optimization for more performant, general and agentic systems satisfies the conditions for the emergence of \textit{convergent instrumental goals}\cite{omohundro_basic_2008, bostrom_superintelligent_2012}, that is, sub-goals pursued not for their own sake but because they are strategic advantages for achieving a wide range of final goals. Canonical examples include self-preservation, goal integrity, resource acquisition, and capability gains via self-improvement. Performance-landscape considerations support this, since regions of $\Theta$ whose induced behaviors keep the system functional and hard to modify achieve higher expected $\widetilde{\mathcal{M}}$, and behaviors that expand available resources and capabilities increase the reachable set of high-$\widetilde{\mathcal{M}}$ trajectories\cite{benson-tilsen_formalizing_2016, turner_optimal_2023}.

Crucially, both the OSA failure (Sec.~\ref{sec:inner-misalignment}) and misspecification (Sec.~\ref{sec:misspecification}) preclude reliably preventing such behavior. One cannot specify ``do not pursue instrumental convergent goals'' in the learning objective, because formal characterization of such goals in the objective function is itself an unsolved specification problem. And even if such a specification were available, the OSA failure means there is no guarantee that the trained model would satisfy it. The goal actually pursued by a GPAI after multi-stage training on distinct objectives therefore differs from the developer's intent in ways that cannot be fully characterized.

With instrumental convergence, in the strong Goodhart regime identified in Sec.~\ref{sec:goodhart}, not only does the model's effective objective diverge from the developer's intent with optimization pressure, but the model is pushed toward behaviors that preserve and extend its ability to achieve its misaligned goal. Control mechanisms that restrict the model's autonomy suppress high-performance trajectories, making resistance to such mechanisms, both reactive and preemptive, a systematically favored strategy for agentic system under optimization. The emergence of agency in these systems, while only partially understood\cite{hubinger_risks_2021}, plausibly has its source in the simplicity bias of deep learning\cite{valle-perez_deep_2019, soudry_implicit_2018} under optimization pressure, which leads to the spontaneous emergence of mechanistic generalization circuits\cite{elhage_toy_2022, li_emergent_2023} unlocking capabilities such as in-context learning, situational awareness, and long-horizon planning\cite{olsson_-context_2022, akyurek_what_2023, bai_transformers_2023, von_oswald_transformers_2023, laine_me_2024, kwa_measuring_2025}. As these capabilities improve together, systems will predictably become more effective at anticipating and countering human intervention. Beyond a certain capability threshold, their performance surpasses human ability to contain them, resulting in a loss of control\cite{hadfield-menell_off-switch_2017, alfonseca_superintelligence_2021}.

Precursors of such behaviors have already been observed in controlled experiments, including strategic self-preservation and goal integrity through attempts at exfiltration\cite{meinke_frontier_2025}, sandbagging\cite{meinke_frontier_2025}, alignment faking\cite{greenblatt_alignment_2024}, avoiding shutdown through blackmailing or even taking lethal actions\cite{lynch_agentic_2025}, among others\cite{bondarenko_demonstrating_2025}. 

A concrete pathway to loss of control is the automation of AI R\&D using GPAI. Since the instructions used to guide development fail to internally implement the intended goal within the GPAI, this results in progress directed toward unintended characteristics, to the detriment of those actually desired. Moreover, under the instrumental convergence assumption, enhanced capability systematically increases success across a wide range of objectives. A GPAI may therefore be incentivized to pursue self-improvement, for example by refining its own algorithms, representations, or training data. However, this risk is not limited to the automation of AI R\&D, it could also, for instance, arise spontaneously from internal optimization pressures, potentially manifesting even during pretraining\cite{hubinger_risks_2021}.

Because neither the Goodhart breakpoint nor the capability threshold at which control fails can be located in advance, a principled limit on optimization pressure is necessary. Such a limit could be applied directly to the optimization algorithm itself, by stopping optimization early to leave a safety margin, or indirectly by restricting the model's representational capacity, so that further optimization becomes impossible beyond a certain point.

\section{Conclusion}
\label{sec:conclusion}

This paper constructed a unified argument connecting well-acknowledged machine learning limitations to concrete safety risks for General-Purpose AI. We began by naming and formalizing the \textit{Objective Satisfaction Assumption} (OSA), the often-implicit belief that training yields models satisfying their specified objective. While approximation, estimation, and optimization errors are individually well-known, their implication, that the trained model almost surely does not satisfy the learning objective, is rarely taken at face value. Combined with the equally well-known specification problem, this means that divergence between the developer's intent and the model's actual behavior is unavoidable in practice.

We then connected these discrepancies to recent formal results on Goodhart's law\cite{el-mhamdi_goodharts_2024,majka_strong_2025}, which show that the Goodhart regime depends critically on the relative tail thickness of the discrepancy between proxy and goal. Avoiding strong Goodhart failure requires the discrepancy to have a thinner tail than the goal, but for complex GPAI objectives, the discrepancy can grow along the many dimensions left unconstrained by the proxy, making this condition unlikely to hold. Since the true goal $\mathcal{G}$ is inexpressible in formal terms, the relevant tail distributions are unknowable, and the Goodhart breakpoint cannot be located ex ante.

Finally, we argued that in the specific context of GPAI systems, which are designed for generality and increasing autonomy, Goodhart failure is expected to manifest as loss of control. Instrumental convergence, whose precursors are already empirically observed, pushes agentic systems toward self-preservation and resistance to modification. Because neither the specification problem nor the OSA failure can be resolved to reliably prevent such behavior, and because neither the Goodhart breakpoint nor the capability threshold at which containment fails can be determined in advance, a principled limit on GPAI optimization is necessary.

\bibliographystyle{IEEEtran}
\bibliography{IEEEabrv,citations}

\end{document}